\documentclass[letterpaper, 10 pt, conference]{ieeeconf}  % Comment this line out if you need a4paper

\IEEEoverridecommandlockouts                              % This command is only needed if 
\usepackage{times}

\usepackage[ruled,vlined]{algorithm2e}
\usepackage[noend]{algcompatible}

\usepackage{amsmath}
\usepackage{amsfonts}
\usepackage{breqn}
\usepackage{multirow,hhline,graphicx,array}
\graphicspath{ {./images/} }
\usepackage{multicol}
\usepackage[bookmarks=true]{hyperref}
\usepackage{subfiles}
\usepackage{caption}
\usepackage{footnote}
\usepackage{gensymb}
\usepackage{float}
\usepackage[flushleft]{threeparttable}

\usepackage{xcolor}

\title{\LARGE \bf
Calibrating LiDAR and Camera using Semantic Mutual information
}

\author{Peng Jiang$^{1}$, Philip Osteen$^{2}$, and Srikanth Saripalli$^{1}$
\thanks{$^{1}$J. Mike Walker '66 Department of Mechanical Engineering, 
Texas A\&M University, College Station, TX 77843, USA
        {\tt\small \{maskjp, ssaripalli\}@tamu.edu}}%
\thanks{$^{2}$DEVCOM Army Research Laboratory (ARL), Adelphi, MD 20783, USA
        {\tt\small philip.r.osteen.civ@mail.mil}}%
}

\begin{document}

\maketitle
\thispagestyle{empty}
\pagestyle{empty}

%%%%%%%%%%%%%%%%%%%%%%%%%%%%%%%%%%%%%%%%%%%%%%%%%%%%%%%%%%%%%%%%%%%%%%%%%%%%%%%%
\begin{abstract}
We propose an algorithm for automatic, targetless, extrinsic calibration of a LiDAR and camera system using semantic information.  We achieve this goal by maximizing mutual information (MI) of semantic information between sensors, leveraging a neural network to estimate semantic mutual information, and matrix exponential for calibration computation. Using kernel-based sampling to sample data from camera measurement based on LiDAR projected points, we formulate the problem as a novel differentiable objective function which supports the use of gradient-based optimization methods. We also introduce an initial calibration method using 2D MI-based image registration. Finally, we demonstrate the robustness of our method and quantitatively analyze the accuracy on a synthetic dataset and also evaluate our algorithm qualitatively on KITTI360 and RELLIS-3D benchmark datasets, showing improvement over recent comparable approaches. %\url{https://github.com/unmannedlab/deep_calibration}.
\end{abstract}

%%%%%%%%%%%%%%%%%%%%%%%%%%%%%%%%%%%%%%%%%%%%%%%%%%%%%%%%%%%%%%%%%%%%%%%%%%%%%%%%
\section{Introduction}
Camera and Light Detection and Ranging (LiDAR) sensors are essential components of autonomous vehicles with complementary properties. A camera can provide high-resolution color information but is sensitive to illumination and lack of spatial data. A LiDAR can provide accurate spatial information at longer ranges and is robust to illumination changes, but its resolution is much lower than the camera and doesn't measure color. Fusing the measurements of these two sensors allows autonomous vehicles to have an improved understanding of the environment. In order to combine the data from different sensor modalities, it's essential to have an accurate transformation between the coordinate systems of the sensors. Therefore, calibration is a crucial first step for multi-modal sensor fusion.

In recent years, many LiDAR-camera calibration methods have been proposed. These methods can be categorized based on whether they are online \cite{Nagy2019}\cite{Wang2020}\cite{Zhu2020}\cite{Lv2020}\cite{Yuan2020}\cite{Chien2016} or offline \cite{Mishra2020},\cite{Mishra2020a} \cite{Owens2015} as well as whether they require a calibration target \cite{Mishra2020},\cite{Mishra2020a} or not \cite{Lv2020}\cite{Yuan2020}\cite{Chien2016}\cite{Zhao2021}. Target-based methods require carefully designed targets \cite{Zhang2004}, \cite{Pusztai2017}, \cite{Owens2015} while targetless methods use data from the natural environment to perform calibration. In this paper, we develop an effective targetless method. 

Calibration algorithms can also be categorized into whether they are optimization-based methods or learning-based methods. Most traditional methods are optimization-based methods that try to minimize or maximize a metric \cite{Mishra2020},\cite{Mishra2020a},\cite{Kang2020},\cite{Taylor2015},\cite{Pusztai2017},]\cite{Zhao2021}. Meanwhile, learning-based methods construct and train neural network models to directly predict the calibration parameters \cite{Lv2020},\cite{Schneider2017},\cite{Zhao2021},\cite{Yuan2020}. Both target-based and targetless optimization-based methods must define features for calibration. Common features include edges\cite{Mishra2020},\cite{Mishra2020a},\cite{Kang2020},\cite{Iyer2018}, gradient\cite{Taylor2015}, and semantic information \cite{Nagy2019,Zhu2020,Wang2020}. Among these, semantic information is a higher-level feature that is available from human annotations or learning-based semantic segmentation model predictions \cite{Zhang2019}, \cite{Yu2018}.
Several methods have been proposed to utilize semantic information to calibrate LiDAR and camera. Nagy et al. \cite{Nagy2019} use a structure from motion (SfM) pipeline to generate points from image and register with LiDAR to create basis calibration and refine based on semantic information. Zhu et al. \cite{Zhu2020} use semantic masks of image and construct height map to encourage laser points to fall on the pixels labeled as obstacles. The methods mentioned above only use semantic information from images and also define other features to help calibration. Wang et al. \cite{Wang2020} proposed a new metric to calibrate LiDAR and camera by reducing the distance between the misaligned points and pixels based on semantic information from both image and point cloud.

In this paper, we also use semantic information from LiDAR and camera measurements. We propose to use mutual information as the metric to optimize. Mutual information is widely used in medical image registration \cite{Oliveira2014, Nan2020}. Pandey et al.\cite{Pandey2015} first proposed to apply it to LiDAR camera calibration. Their approach considers the sensor-measured surface intensities (reflectivity for LiDAR and grayscale intensity for camera) as two random variables and maximizes the mutual information between them. Inspired by \cite{Nan2020}, we use mutual information neural estimate (MINE) \cite{Belghazi2018} to estimate the mutual information. We also use matrix exponential to compute transformation matrix and use the kernel-based sampler to sample points from images based on projected LiDAR. These treatments allow us to propose a fully differentiable LiDAR camera calibration framework based on semantic information, and implement the algorithm using popular Deep learning libraries.
The major contributions of this dataset can be summarized as follows:
\begin{itemize}
  \item We propose an algorithm for automatic, targetless, extrinsic calibration of a LiDAR and camera system using semantic information.
  \item We show how to make the objective function fully differentiable and optimize it using the gradient-descent method. 
  \item We introduce an initial calibration method using 2D MI-based image registration. 
  \item We evaluate our method on a synthetic dataset from Carla simulator \cite{Dosovitskiy17}, as well as the real KITTI360 \cite{Xie2016CVPR} and RELLIS-3D \cite{jiang2020rellis3d} datasets.
\end{itemize}
\section{Methology}
\subsection{Algorithm}
\begin{figure*}
  \centering
  \includegraphics[width=\textwidth]{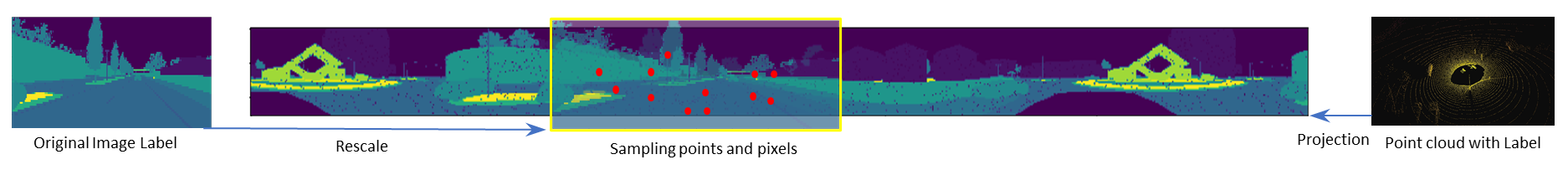}
  \caption{Initial Calibration Procedure: (a) Project LiDAR label into 2D cylinder plane and zoom the camera image label. (b) Register 2D semantic Images. (c) Sample points and pixels from the overlapped area of the two semantic labels.}
  \label{fig:initial}
\end{figure*}
\begin{algorithm}
\SetAlgoLined
\SetKwData{Left}{left}\SetKwData{This}{this}\SetKwData{Up}{up}
\SetKwData{Repeat}{repeat}\SetKwData{Until}{Until}
\SetKwFunction{Union}{Union}\SetKwFunction{FindCompress}{FindCompress}
\SetKwInOut{Input}{input}\SetKwInOut{Output}{output}
\Input{Lidar Point Cloud $P$, Point Cloud Labels $L^{P}$, Image Labels $L^{C}$, Camera Intrinsic Matrix $K$, Initial Transformation Matrix $T_{init}$}
\Output{Transformation Matrix $T$}
 Use $T_{init}$ to initialize $v_i$\; 
 Use random initialization for MINEnet parameters $\theta$\;
 Initialize learning rate $\alpha$, $\beta$, optimizer and learning rates scheduler\;
 \While{not converge}{
    $T = \sum_{i=0}^{5} {v_iB_i}$\;
    Sample $b$ minibatch points $P_{b}$ and labels $L_{b}^{P}$ from $P$ and $L^{P}$\;
    $P_{b}^{u,v}, = \text{Proj}(P_{b},T,K)$\;
    $\widetilde{L}_{b}^{C} = \text{Sample}(L^{C},P_{b}^{u,v})$\;
    $MI = \text{MINE}(L_{b}^{P},\widetilde{L}_{b}^{C}$)\;
    Update MIME parameter: $\theta += \alpha\bigtriangledown_{\theta}\text{MI}$\;
    Update matrix exponential parameters: $v += \beta\bigtriangledown_{v}\text{MI}$\;
 }
 Return $T = \sum_{i=0}^{5} {v_iB_i}$\;
\caption{3D Calibration}
\label{alg:3d}
\end{algorithm}

The following formula describes the transformation relationship of a point $p^L_i$ from the LiDAR local coordinate system to the camera projection plane:
\begin{equation}
\begin{bmatrix}p^{uv}_{i}\\1\end{bmatrix}=\mathrm{K}\begin{bmatrix}
1 & 0 & 0 & 0 \\
0 & 1 & 0 & 0 \\
0 & 0 & 1 & 0
\end{bmatrix}\begin{bmatrix}
\mathrm{R} & \mathrm{t} \\
0 & 1
\end{bmatrix} \begin{bmatrix}p^L_i\\1\end{bmatrix}
\label{eq:lidarcam}
\end{equation}

\begin{itemize}
    \item $p_i^L=\begin{bmatrix}x_{i}^L&y_{i}^L&z_{i}^L\end{bmatrix}^T$ represents the coordinate of point $p_i$ in Lidar local frame;
    %\item $t = \begin{bmatrix}t_{x}&y_{y}&t_{z}\end{bmatrix}^T$ represents translation vector;
    \item $t$ represents the $3\times1$ translation vector;
    \item $R$ represents the $3\times3$ rotation matrix;
    %\item $R =  \begin{bmatrix}r_{11}&r_{12}&r_{13}\\r_{21}&r_{22}&r_{23}\\r_{31}&r_{32}&r_{33}\end{bmatrix} $ represents rotation matrix;  
    %\item $K =   \begin{bmatrix}f_{x}&0&c_{x}\\0&f_{y}&c_{y}\\0&0&1\end{bmatrix}  $ represents intrinsic calibration matrix of camera;  
    \item $K$ represents the $3\times3$ intrinsic matrix of camera;  
    %\item $p_i^C=\begin{bmatrix}x_{i}^C&y_{i}^C&z_{i}^C\end{bmatrix}^T$ represents the coordinate of point $p_i$ in Camera local frame;
    \item $p_i^{uv}=\begin{bmatrix}u_{i}&v_{i}\end{bmatrix}^T$ represents the coordinate of point $p_i$ in Camera projection plane;
\end{itemize}
This paper focuses on the extrinsic calibration between LiDAR and camera; therefore, we assume that the camera and LiDAR have been intrinsically calibrated. The extrinsic calibration parameters are given by $R$ and $t$ in Eq.\ref{eq:lidarcam}. In addition to point cloud coordinates and image, we also assume the semantic labels of point cloud $L^L$ and image $L^C$ are available. 

We consider the semantic label value of each point cloud and its corresponding image pixel as two random variables $X$ and $Y$. The mutual information of the two variable have the maximum value when we have the correct calibration parameters. In order to perform calibration, we  need to perform the following three operations:
\begin{enumerate}
    \item \textbf{Transformation Matrix Computation}: $P^{uv} = \text{Proj}(P^{L},R,t)$ projects point cloud from lidar coordinate to camera coordinate;
    \item \textbf{Image Samping}: $\widetilde{L}^{C} = \text{Sample}(L^{C},P^{uv})$ samples semantic label values from image labels based on projected LiDAR coordinates.
    \item \textbf{Mutual Information Estimation}: $I(X,Y) = \text{MI}(\widetilde{L}^{C},\widetilde{L}^{L})$ estimate mutual information base the samples from the semantic labels of LiDAR points and its corresponding pixels on image.
\end{enumerate}
 Therefore, our full optimization objective function can be written as Eq.\ref{eq:opt1}
%\begin{equation}
%    \hat{T}=\arg \max _{T}I(X,Y; T)
%\label{eq:opt1}
%\end{equation}
\begin{equation}
    R,t=\arg \max _{R,t}\text{MI}(\text{Sample}(L^{C},\text{Proj}(P^{L},R,t)))
\label{eq:opt1}
\end{equation}
\subsection{Optimization}
The cost function Eq.\ref{eq:opt1} is maximized at the correct value of the rigid-body transformation parameters between LiDAR and camera. Therefore, any optimization technique that iteratively converges to the global optimum can be used, although we prefer gradient based optimization methods for their fast convergence properties. Here, we present how to make the cost function fully differentiable, which allows us to optimize Eq.\ref{eq:opt1} with a gradient-based optimization method. In the remaining part of this section, we describe how to make transformation matrix computation, image sampling and mutual information estimation differentiable, with the full algorithm given in Algorithm.\ref{alg:3d}.
\subsubsection{Transformation Matrix Computation}
$P^{uv} = \text{Proj}(P^{L},R,t)$ involves rigid 3D transformation $T$ which can be represented as the matrix exponential ($ T = \exp{(H)}$). And the matrix $H$ can be parameterized as a sum of weighted basis matrix of the Lie algebra $se(3)$ ($ H = \sum_{i=0}^{5} {v_iB_i}$) as described in \cite{Wachinger2013}. Therefore, we can represent $T$ with 6 parameters (see Eq.\ref{eq:expT}) and apply standard partial derivative to compute the gradient while optimizing. 
\begin{equation}
T  = \begin{bmatrix}R&t\\0&1\end{bmatrix} =\exp{(\sum_{i=0}^{5} {v_iB_i})}= \sum_{j=0}^{\infty } { \frac{(\sum_{i=0}^{5} {v_iB_i})^n }{n!} }
\label{eq:expT}
\end{equation}

\subsubsection{Image Sampling}
Differentiable image sampling is widely used in deep learning for computer vision \cite{Jaderberg2015}. The sampling operation can be written as
\begin{equation}
    \widetilde{l}_{i}^{C}=\sum_{h}^{H} \sum_{w}^{W} l_{h w}^{C} k\left(u_{i}-h; \Phi_{x}\right) k\left(v_{i}-w ; \Phi_{y}\right) 
\end{equation}
where $\Phi_{x}$ and $\Phi_{y}$ are the parameters of a generic sampling kernel $k()$ which defines the image interpolation (e.g. bilinear), $l_{h w}^{C}$ is the label value at location $(h,w)$ of the image semantic label. $\widetilde{l}_{i}^{C}$ is the corresponding semantic label of point $p_i$ after projecting on image plane.
\subsubsection{Mutual Information Estimation}
Mutual information is a fundamental quantity for measuring the relationship between random variables. Traditional approaches are non-parametric (e.g., binning, likelihood-ratio estimators based on support vector machines, non-parametric kernel-density estimators), which are not differentiable. Several mutual information neural estimators have been proposed \cite{Belghazi2018, Mukherjee2019, KumarMondal}. These estimators are consist of neural networks that are fully differentiable. In our implementation,  we use MINE\cite{Belghazi2018} to estimate mutual information. This method uses the Donsker-Varadhan (DV) duality to represent MI as
\begin{equation}
 \widehat{I(X ; Y)}_{n}=\sup _{\theta \in \Theta} \mathbb{E}_{P_{(X,Y)}}\left[F_{\theta}\right]-\log \left(\mathbb{E}_{P(X) P(Y)}\left[e^{F_{\theta}}\right]\right)
\label{eq:mime}
\end{equation},
where $P(X,Y)$ is the joint density for random variables $X$ and $Y$ and $P(X)$ and $P(Z)$ are marginal densities for $X$ and $Y$. $F_{\theta}$ is a function parameterized by a neural network, and $\theta$ is the parameters of the neural network.

\subsection{Initial Calibration}\label{sec:initial}
Most targetless calibration approaches require good initialization because good initialization helps the optimization converge faster. Besides, good initialization can help gradient-based optimization methods to avoid some local minimum. By utilizing the semantic information, we convert the initial calibration method into a combination of 2D image registration and Perspective-n-Point
(PnP) problem\cite{Lepetit2009}. We project LiDAR points into a spherical 2D plane \cite{Jiang2020} and get a 2D semantic range image from the LiDAR. We consider the LiDAR to be a low-resolution 360-degree camera with a partially overlapping field of view with the ordinary camera as shown in Fig.\ref{fig:initial}. Therefore, we zoom the semantic label of the camera image into the same resolution as LiDAR. Then, we perform 2D MI-based image registration on the two semantic images. After registration, we can get the raw correspondence between points of LiDAR and pixels of the camera image. Then, we sample several pairs of points and pixels and solve the PnP problem to get an initial calibration between the LiDAR and camera. The full algorithm is described in Algorithm.\ref{alg:init}

\begin{algorithm}
\SetAlgoLined
\SetKwData{Left}{left}\SetKwData{This}{this}\SetKwData{Up}{up}
\SetKwData{Repeat}{repeat}\SetKwData{Until}{Until}
\SetKwFunction{Union}{Union}\SetKwFunction{FindCompress}{FindCompress}
\SetKwInOut{Input}{input}\SetKwInOut{Output}{output}
\Input{Lidar Point Cloud $P$, Point Cloud Labels $L^{P}$, Lidar Field of View $FoV_{H}^{L},FoV_{V}^{L}$, Lidar Channel Number $H^{L}$ and Ring Point Number $W^{L}$  Image Labels $L^{I}$,  Camera Field of View $FoV_{H}^{C}, FoV_{V}^{C}$, Image Height $H^{I}$ and Width $W^{I}$ }
\Output{Initial Transformation Matrix $T_{init}$}
$L^{P}_{cy} = \text{SphericalProj}(P,L^{P},H^{L},W^{L})$\;
$W_{z}^{I},H_{z}^{I} =  \frac{W^{L}}{Fov_{V}^{L}}FoV_{V}^{C}, \frac{H^{L}}{FoV_{H}^{L}}Fov_{H}^{C}$\;

$L^{I}_{z} = \text{Zoom}(L^{I},W_{z}^{I},H_{z}^{I})$\;
Register $L^{I}_{z}$ and $L^{P}_{cy}$ using 2D MI-based method\;
Sample pixels $I^{P}_{cy}$ and $I^{I}_{z}$ from the overlapping between $L^{P}_{cy}$ and $L^{I}_{z}$\;
Recover image pixels $I^{I}_{s} = \text{DeZoom}(I^{I}_{z},W_{z}^{I},H_{z}^{I},W^{I},H^{I})$\;
Recover points $P_{s} = \text{DeSphericalProj}(I^{P}_{cy},P)$\;
$T_{init} = \text{PnPsolver}(P_{s},I^{I}_{s})$\;
\caption{Initial Calibration}
\label{alg:init}
\end{algorithm}

\begin{figure}
  \centering
  \includegraphics[width=0.5\textwidth]{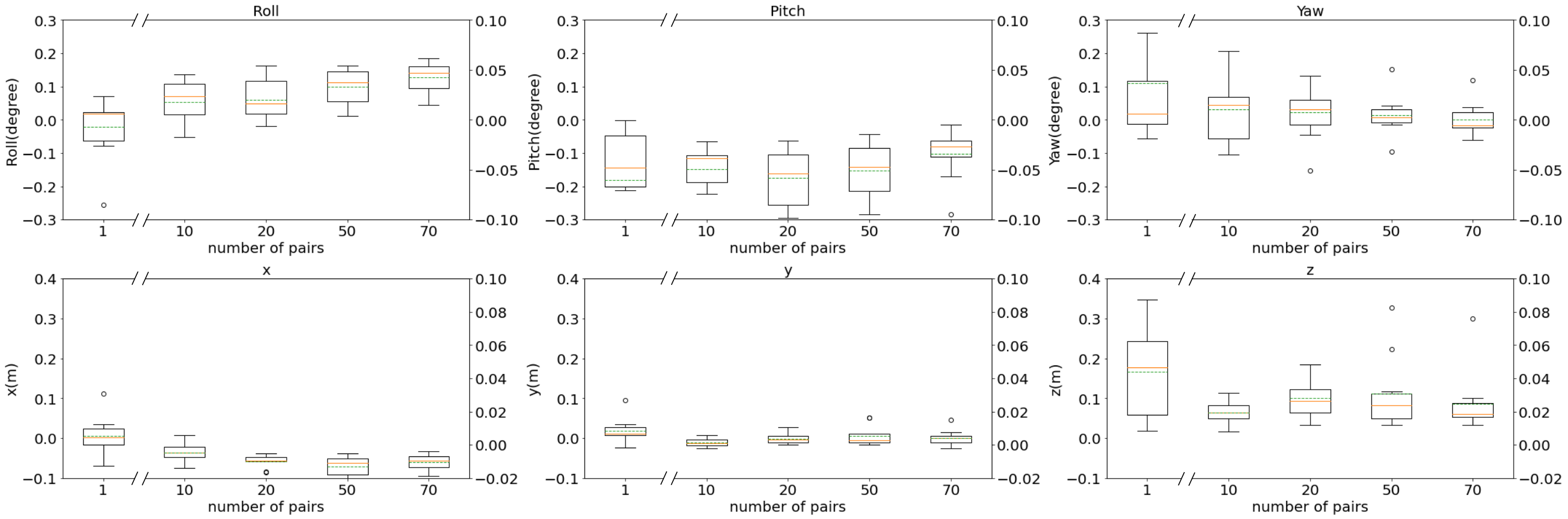}
  \caption{Calibration error using different number of pairs.}   
  \label{fig:boxplot}
\end{figure}
\vspace{-10pt}
\section{Experiment and Results}
This section describes the experiments to evaluate the accuracy and robustness of the proposed automatic calibration technique. We first evaluate our methods on a synthetic dataset. Then we tested our methods on the real-world KITTI360 \cite{Xie2016CVPR} and RELLIS-3D \cite{jiang2020rellis3d} datasets. 
\subsection{Synthetic dataset}
\begin{figure}
  \centering
  \includegraphics[width=0.45\textwidth]{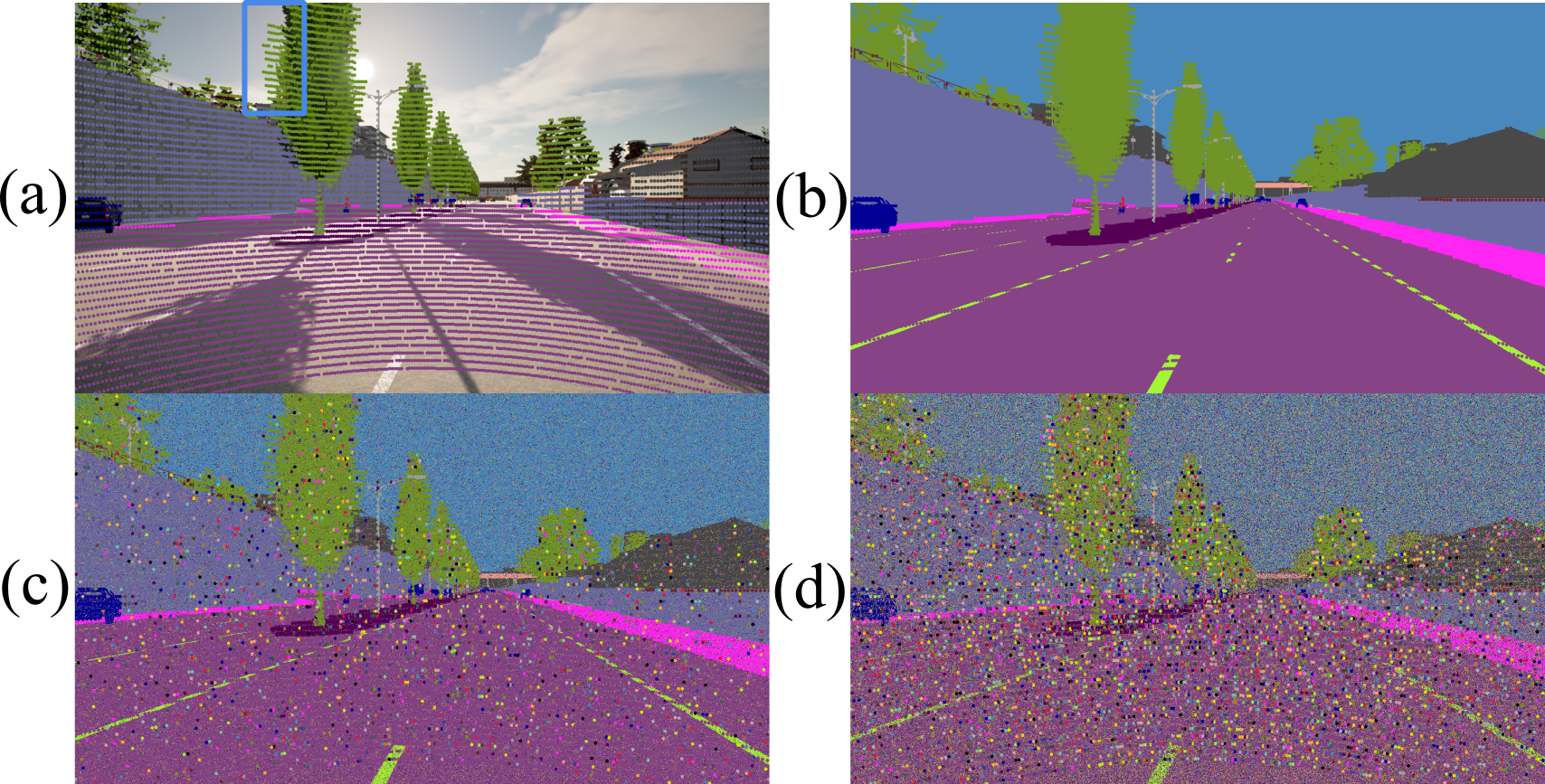}
  \caption{(a) Simulated LiDAR point cloud projected on image; (b) Ground true image label with point cloud; (c) image label and point cloud label with 20\% error (d)  image label and point cloud label with 50\% error.}   
  \label{fig:noise_label}
\end{figure}

To test the accuracy and robustness of our methods, we created a dataset including paired LiDAR and camera data with semantic labels using the Carla simulator \cite{Dosovitskiy17}. The simulator can generate 21 classes. During, experiments we only used 20 classes. The image has a size of $1280\times720$. The Lidar has 360 degrees and 64 channels. Each channel includes roughly 800 points per frame. Due to the limitation of simulation, the point cloud and image are completely overlapped with each other (see Fig.\ref{fig:noise_label} (a)). However, the synthetic data can still provide the most accurate transformation and semantic labels. Another disadvantage of synthetic data is that some real-world sensor and environmental characteristics are not perfectly modeled.

\begin{figure}
  \centering
  \includegraphics[width=0.5\textwidth]{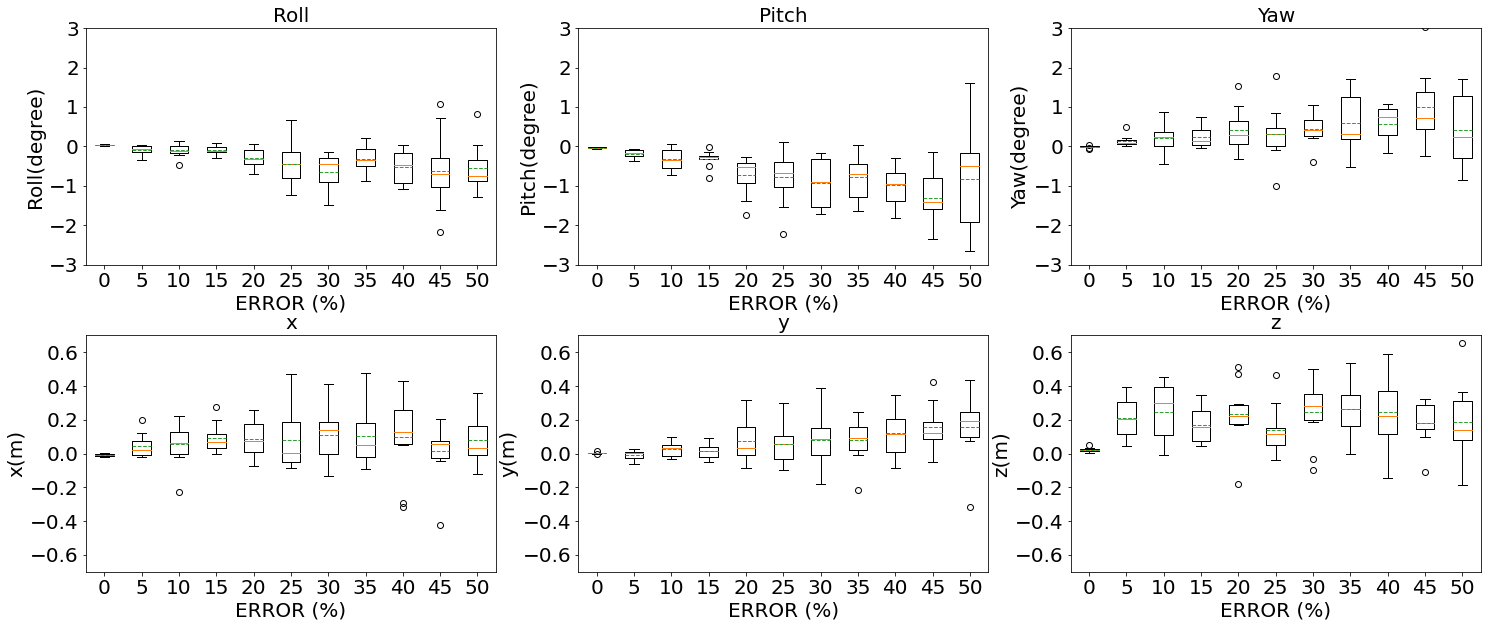}
  \caption{Calibration error of different noisy labels}   
  \label{fig:boxplot_noise}
 \vspace{-10pt}
\end{figure}

We first tested the performance and robustness of our methods and the effects of the number of data frames on overall performance. In each test, we used the ground truth labels and provided the initial transformation through the methods described in section \ref{sec:initial}. Then, we tested the procedure with 1,10,20,50, and 70 pairs of frames. As shown in Fig.\ref{fig:boxplot}, the variance of the error decreases as we increase the number of pairs we use. Because by introducing more data, the estimator is able to have better mutual information estimation of the overall semantic information. Besides, more data also increase the scene's diversity and reduce the local minimum of the optimization. 
Secondly, we tested the effects of the error of labels. In each test, we added random noise to both the image label and point cloud label (see Fig.\ref{fig:noise_label}). The results are shown in Fig.\ref{fig:boxplot_noise}. As expected, the accuracy decreases, and the variance increases with the error of the labels.
\begin{table}[]
\vspace{-10pt}
\caption{Calibration results of KITTI360}
\centering
\begin{tabular}{  c | c | c | c | c | c | c } 
\hline
Methods &  Roll($^{\circ}$) &  Pitch($^{\circ}$) & Yaw($^{\circ}$) & X(m) & Y(m)  & Z(m) \\ 
\hline
KITTI360 &  73.72&  -71.37&64.56 & 0.26& -0.11  &  -0.83 \\ 
\hline
SOIC & 74.08 &  -71.21 &  64.75 & 0.11 & -0.12 &  -1.29 \\ 
\hline
PMI &  71.19 & --64.38 &  58.35 & 2.28 & -0.01 & -0.70\\ 
\hline
Ours &  73.98 & -71.18 &  64.57 & 0.02 & -0.14 & -1.36 \\ 
\hline
\end{tabular}
    \label{tab:kitti360}
\end{table}

\begin{table}[]
    \centering
\caption{Calibration results of RELLIS3D}
    \begin{tabular}{ c | c | c | c | c | c | c } 
    \hline
Methods &  Roll($^{\circ}$) &  Pitch($^{\circ}$) & Yaw($^{\circ}$) & X(m) & Y(m)  & Z(m) \\ 
    \hline
    RELLIS3D &  70.80 &  67.78 & -68.06 & -0.04 & -0.17 & -0.13\\ 
    \hline
    SOIC & 70.08 &   68.62 & -66.68 & 0.00 & -0.15 &  4.41 \\ 
    \hline
    PMI & 75.65 &  69.87 & -64.24 & -0.08 &  -0.33 & -0.10 \\ 
    \hline
    Ours &  70.24&  67.63 & -67.32 & -0.05 & -0.19 & -0.06 \\ 
    \hline
    \end{tabular}
    \label{tab:rellis3d}
\end{table}

\begin{figure*}
  \centering
  \includegraphics[width=\textwidth]{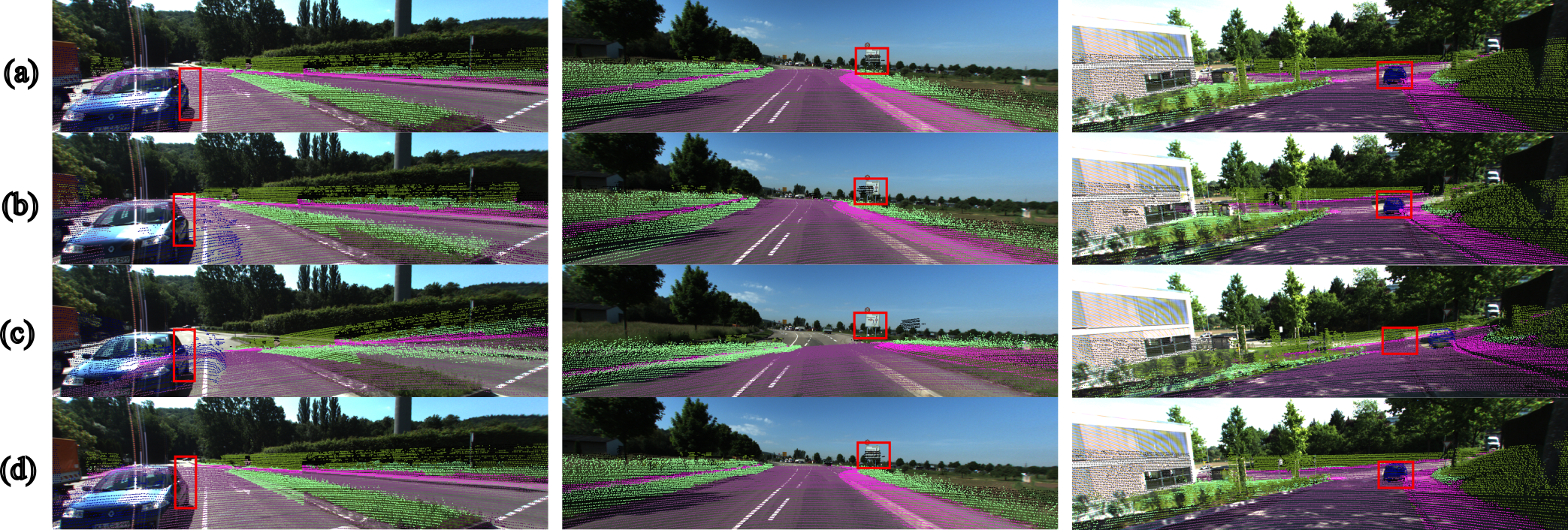}
  \caption{(a) KITT360 calibration; (b) SOIC Results; (c) MI using surface intensities; (d) MI using semantic information. }
  \label{fig:kitti360}
\end{figure*}
\begin{figure*}
  \centering
  \includegraphics[width=\textwidth]{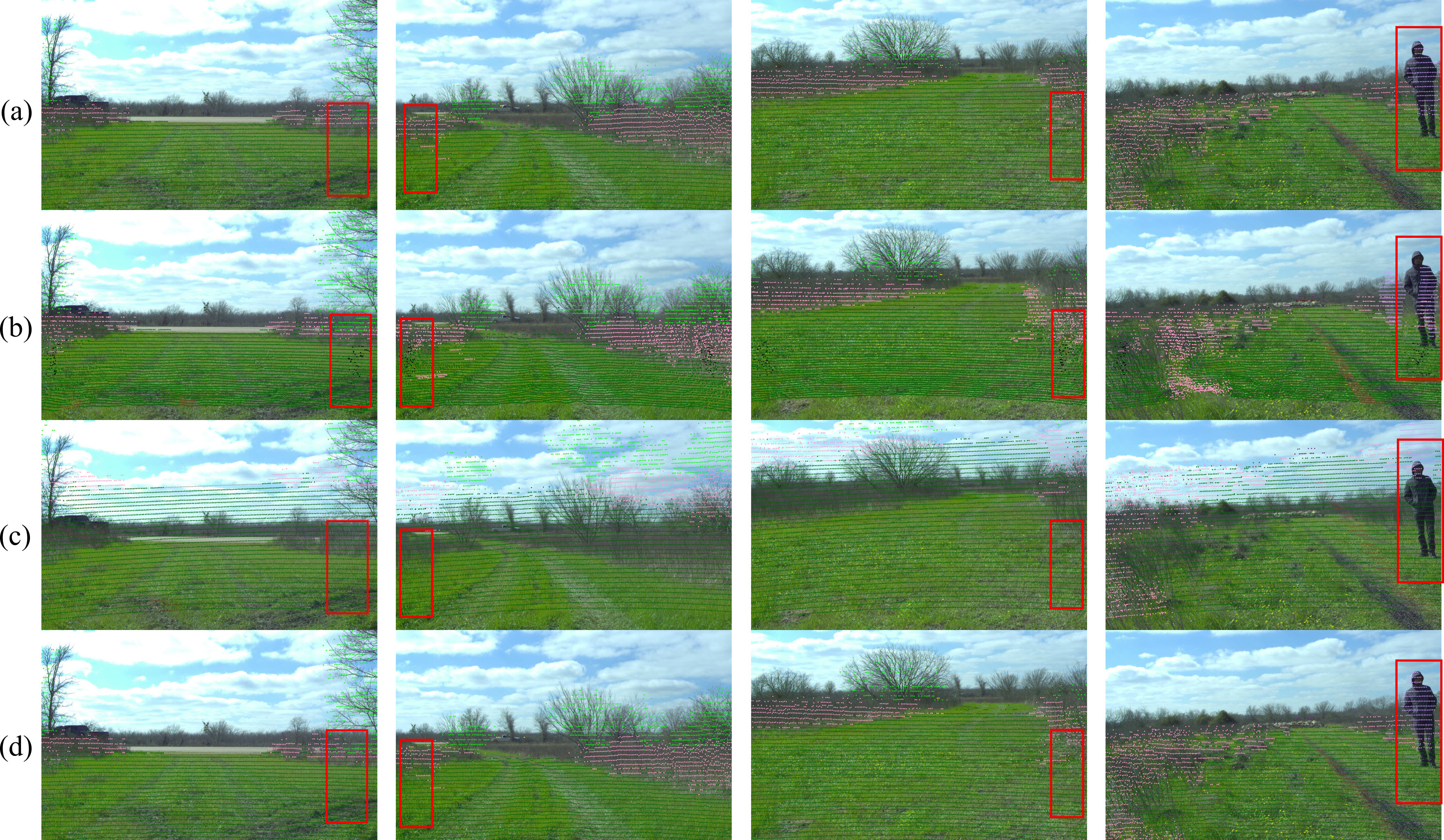}
   \caption{(a) RELLIS-3D calibration; (b) SOIC Results; (c) MI using surface intensities; (d) MI using semantic information. }
  \label{fig:rellis3d}
\end{figure*}
\subsection{Real-world datasets}
Following our experiments with simulated data, we tested on two real-world datasets, KITTI360 and RELLIS-3D, which both provide annotated synchronized LiDAR and image data. Both datasets also provide calibration parameters between LiDAR and camera but they are not accurate as we can see in Fig.\ref{fig:kitti360} (a) and Fig.\ref{fig:rellis3d} (a). Therefore, we apply our methods on the two datasets and qualitatively compared with SOIC\cite{Wang2020}. We also follow Pandey's \cite{Pandey2015} method to use the mutual information between the sensor-measured surface intensities to calibrate the sensors. But we use MINE\cite{Belghazi2018} to estimate the MI.The results is noted as \textbf{PMI} in Table\ref{tab:kitti360} and \ref{tab:rellis3d}.All methods were initialized by our proposed method.  As shown in Table. \ref{tab:kitti360}-\ref{tab:rellis3d}, our method provide the closest results with the provided parameters of each dataset. Meanwhile, Fig. \ref{fig:kitti360} and \ref{fig:rellis3d} shows that our calibration results better cross-sensor data fusion than the other two methods. More visual results can be found in the video\footnote{\url{https://youtu.be/nSNBxpCtMeo}}.

\section{Summary and Future Work}
This paper presented a fully differential LiDAR-Camera calibration method using semantic information from both sensor measurements. By utilizing semantic information, the method doesn't need specific targets and initial guess parameters. 
Because the method is fully differentiable, it can be implemented using the popular deep learning framework, such as Pytorch\cite{Paszke2019} or Tensorflow\cite{Abadi2016}. Moreover, mutual semantic information was introduced used to register multi-modal data. The method has the potential to leverage deep features to calibrate LiDAR and camera. Another possible application of this method is to use this framework in deep fusion directly. By embedding this framework in the deep fusion framework, rough calibration between sensors might be enough.

%\section*{APPENDIX}


%\section*{ACKNOWLEDGMENT}

%%%%%%%%%%%%%%%%%%%%%%%%%%%%%%%%%%%%%%%%%%%%%%%%%%%%%%%%%%%%%%%%%%%%%%%%%%%%%%%%
\bibliographystyle{IEEEtran}
\bibliography{IEEEabrv,references}

\end{document}